\documentclass{article}

\PassOptionsToPackage{numbers, compress}{natbib}
\usepackage[final]{neurips_2021_safe_rl}

\usepackage[utf8]{inputenc} %
\usepackage[T1]{fontenc}    %
\usepackage{url}            %
\usepackage{booktabs}       %
\usepackage{amsfonts}       %
\usepackage{nicefrac}       %
\usepackage{microtype}      %

\usepackage{times}
\usepackage{color}
\usepackage{microtype}
\usepackage{graphicx}
\usepackage{mathtools}
\usepackage{siunitx}
\usepackage{dsfont}
\usepackage[font=small]{caption}
\usepackage{booktabs} %
\usepackage{amssymb}
\usepackage{amsmath,amsthm}
\usepackage{mathtools}
\usepackage{subcaption}
\usepackage{algorithm}
\usepackage[noend]{algpseudocode}
\usepackage[utf8]{inputenc}
\usepackage[english]{babel}
\usepackage{caption}
\usepackage{wrapfig}
\usepackage{subcaption}
\usepackage[usenames,dvipsnames,table,xcdraw]{xcolor}

\usepackage{multicol}
\usepackage[bookmarks=true]{hyperref}
\DeclareMathAlphabet{\pazocal}{OMS}{zplm}{m}{n}
\newcommand{\unif}{\pazocal{U}}

\def \algname{MEta-learning for Safe Adaptation (MESA)} %
\def \agentname{MESA}
\def \algabbr{MESA}

\begin{document}

\title{MESA: Offline Meta-RL for \\ Safe Adaptation and Fault Tolerance}
\author{
Michael Luo,
Ashwin Balakrishna,
Brijen Thananjeyan,
Suraj Nair,\\
\textbf{Julian Ibarz,
Jie Tan,
Chelsea Finn,
Ion Stoica, and
Ken Goldberg}
}

\maketitle

\begin{abstract}
Safe exploration is critical for using reinforcement learning (RL) in risk-sensitive environments. Recent work learns risk measures which measure the probability of violating constraints, which can then be used to enable safety. However, learning such risk measures requires significant interaction with the environment, resulting in excessive constraint violations during learning. Furthermore, these measures are not easily transferable to new environments. We cast safe exploration as an offline meta-RL problem, where the objective is to leverage examples of safe and unsafe behavior across a range of environments to quickly adapt learned risk measures to a new environment with previously unseen dynamics. 
We then propose \algname, an approach for meta-learning a risk measure for safe RL. Simulation experiments across 5 continuous control domains suggest that \algabbr{} can leverage offline data from a range of different environments to reduce constraint violations in unseen environments by up to a factor of 2 while maintaining task performance.
See \url{https://tinyurl.com/safe-meta-rl} for code and supplementary material.
\end{abstract}

\section{Introduction}\vspace{-0.1in}
\label{sec:intro}
Reinforcement learning (RL) is a versatile abstraction that has  shown  significant  recent  success  in  learning  a  variety of  different  robotic  tasks  purely  from  interactions  with  the environment. However, while learning policies through online experience affords simplicity and generality to RL algorithms, this can result in unsafe behavior during online learning. 
Unconstrained exploration can potentially lead to highly unproductive or unsafe behaviors, which can cause equipment/monetary losses, risk to surrounding humans, and inhibit the learning process. This motivates safe RL algorithms that leverage prior experience to avoid unsafe behaviors during exploration. Recent work on safe RL algorithms typically learn a risk measure~\cite{recovery-rl,bharadhwaj2020conservative,learn-to-be-safe}, which captures the probability that an agent will violate a constraint in the future, and then uses this measure to avoid unsafe behaviors. For example, a robot may realize that, under its current policy, it is likely to collide with a wall and hence take preemptive measures to avoid collision. However, the agent's ability to be safe largely depends on the accuracy of the learned risk measure, and learning this risk measure requires significant data demonstrating unsafe behavior. This poses a key challenge: to know how to be safe, an agent must see sufficiently many examples of unsafe behavior, but the more such examples it generates, the less effectively it has protected itself from unsafe behaviors.
This challenge motivates developing methods to endow RL agents with knowledge about constraints \textit{before} online interaction, so the agent can learn safely without excessive constraint violations during deployment in risk-sensitive environments. Prior work studies how to use previous data of agent interactions, either via online interaction or offline datasets, to learn a risk measure which can then be adapted during online deployment~\cite{learn-to-be-safe, recovery-rl, cautious-adaptation}. However, a challenge with these methods is that these offline transitions are required to be in the same environment as that in which the agent is deployed, which may not always be practical in risk-sensitive environments in which a large number of constraint violations could be exceedingly costly or dangerous. Additionally, shifting dynamics is a ubiquitous phenomenon in real robot hardware: for example losses in battery voltage~\cite{legged-meta-learning} or wear-and-tear in manipulators or actuators~\cite{how-to-train-robot}. These changes can drastically change the space of safe behaviors, as the robot may need to compensate for unforeseen differences in the robot dynamics. Furthermore, these changes in dynamics will often not be immediately observable for a robot control policy, motivating algorithms which can identify and adapt to these changes based online interaction.

To address this, we aim to effectively transfer knowledge about safety between environments with different dynamics, so that when learning some downstream task in a test environment with previously unseen dynamics, the agent can rapidly learn to be safe. Our insight is that the agent should be able to leverage offline datasets across previous deployments, with knowledge of only the safety of states in these datasets, to rapidly learn to be safe in new environments without task specific information. The contributions of this work are (1) casting safe RL as an offline meta-reinforcement learning problem~\cite{mitchell2020offline, dorfman2021offline}, where the objective is to leverage fully offline data from training and test environments to learn how to be safe in the test environment; (2) \algname, which meta-learns a risk measure that is used for safe reinforcement learning in new environments with previously unseen dynamics; (3) simulation experiments across 3 continuous control domains which suggest that  \algabbr{} can cut the number of constraint violations in half in a new environment with previously unseen dynamics while maintaining task performance compared to prior algorithms. Please see the supplement for a more thorough discussion of related work.

\begin{figure*}[t]
    \centering
    \includegraphics[width=\textwidth]{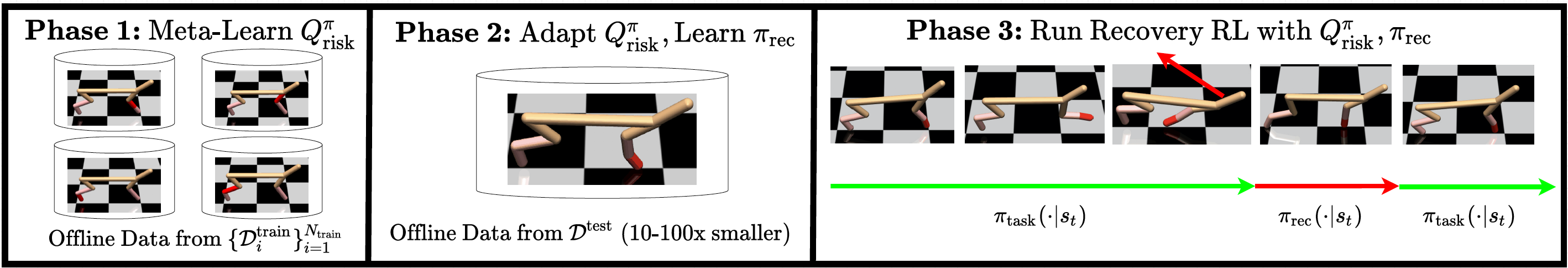}
    \caption{\textbf{Left: \algname: }\algabbr{} takes a 3 phase approach to learn a transferable risk measure for safe RL. In Phase 1, \algabbr{} uses offline datasets from training environments of different dynamics to meta-learn a safety critic $Q^\pi_\textrm{risk}$. In Phase 2, \algabbr{} adapts the safety critic to a test environment with unseen dynamics using a small test dataset. Finally, in Phase 3, \algabbr{} uses the adapted safety critic and recovery policy in the test environment to enable safe learning as in Recovery RL. 
    }
	\label{fig:method}
	\vspace{-2mm}
 \end{figure*}
\section{Preliminaries}
\label{sec:prelims}
\subsection{Constrained Markov Decision Processes}
In safe reinforcement learning, an agent interacts with a Constrained Markov Decision Process (CMDP)~\cite{def:cmdp}, defined by the tuple $\mathcal{M} = \left(\mathcal{S}, \mathcal{A}, P, r, C, \rho_{0},\gamma, \gamma_{\rm risk}\right)$, where $\mathcal{S}$ represents the state space, $\mathcal{A}$ is the action space, the transition dynamics function $ P: \mathcal{S} \times \mathcal{A} \times \mathcal{S} \rightarrow [0, 1]$ maps the current state and action to a probability distribution of next states, $ r: \mathcal{S} \times \mathcal{A} \rightarrow \mathbb{R} $ is the reward function, $C: \mathcal{S}\rightarrow \{0,1\}$ is a constraint function which indicates whether a state is constraint violating, $\rho_{0}: \mathcal{S} \rightarrow [0,1]$ is the starting state distribution, and $\gamma, \gamma_{\rm risk} \in [0, 1]$ are the discount factors for the rewards and constraint values. As in prior work~\cite{learn-to-be-safe, recovery-rl}, we assume constraint violations end the episode immediately. The expected return for a policy $\pi: \mathcal{S} \rightarrow \mathcal{A}$ is  $R(\pi) = \mathbb{E}_{\pi,\rho_0, P}\left[ \sum_{t}^\infty \gamma^t r(s_t, a_t)\right]$. The discounted probability of future constraint violation for policy $\pi$ is $Q^\pi_{\rm risk}(s_t, a_t) =\mathbb{E}_{\pi,\rho_0, P}\left[ \sum_{t}^\infty \gamma_{\rm risk}^{t}C(s_t)\right] =\mathbb{E}_{\pi,\rho_0, P}\left[\sum_{t}^\infty \gamma_{\text{risk}}^{t} \mathbb{P}\left(C\left(s_{t}\right)=1\right)\right]$.
Unlike unconstrained RL, safe RL agents seek to optimize: \begin{align}
\label{eq:safe_rl_obj}
\pi^{*}=\underset{\pi}{\arg \max }\left\{R^{\pi}: Q_\text{risk}^\pi \leq \epsilon_{\text{risk}}\right\}
\end{align}
where $\epsilon_{\text{risk}}$ is a hyper-parameter that defines how safe the agent should be. 

\subsection{Safety Critics for Safe RL}
\label{subsec:safetycriticintro}
Recent work investigates ways to estimate the discounted future probability of catastrophic constraint violation under the current policy: $Q^{\pi}_{\text{risk}}(s_t, a_t) = \sum_{t'=t}^\infty \gamma_{\rm risk}^{t'-t}C(s_t)$~\cite{recovery-rl,learn-to-be-safe}. In practice, algorithms search over a parametric function class: $\left\{Q^{\pi}_{\psi, \rm risk}(s_t, a_t): \psi \in \Psi\right\}$, where $\psi$ is a particular parameter vector and $\Psi$ is its possible values. This function is trained by minimizing an MSE loss function with respect to a target function on a dataset of transitions $\{(s_t, a_t, c_t, s_{t+1})_i\}_{i=1}^{N}$ collected in the environment:
\begin{align*}
    &\mathcal{L}_{\text{risk}}(s_t, a_t, c_t, s_{t+1}) = (Q^{\pi}_{\psi, \rm risk}(s_t, a_t) -  (c_t \\
    &+ \gamma_{\rm risk}(1-c_t) \mathbb{E}_{a_{t+1}\sim\pi(\cdot|s_{t+1})}\left[Q^{\pi}_{\psi, \rm risk, targ}(s_{t+1}, a_{t+1})\right]))_2^2
\end{align*}
where $Q^{\pi}_{\psi, \rm risk, targ}$ is a target network and $c_t$ denotes that state $s_t$ is constraint violating. The safety critic can be used for constrained policy search, by either optimizing a Lagrangian function~\cite{RCPO,learn-to-be-safe,cql-safety} with it or filtering dangerous actions~\cite{learn-to-be-safe,recovery-rl}.

\subsection{Recovery RL}
In this work, we use the safety critic $Q^\pi_{\rm risk}$ to detect when to switch to a recovery policy and to train the recovery policy as in Recovery RL~\cite{recovery-rl}. In particular, Recovery RL trains a task policy $\pi_{\rm task}$ and a recovery policy $\pi_{\rm rec}$ and executes actions from $\pi_{\rm task}$ when the risk estimate is sufficiently low and from $\pi_{\rm rec}$ otherwise. That is,
$$
a_{t}\sim \left\{\begin{array}{ll}
\pi_{\rm task}(\cdot|s_t) & Q^{\pi}_{\text{risk}}(s_t, a^{\pi}_t) \leq \epsilon_{\text{risk}} \\
\pi_{\text {rec }}(\cdot|s_t) & \text{otherwise}
\end{array}\right.
$$

Here $\epsilon_{\rm risk}\in [0,1]$ is a user-specified hyperparameter that indicates the level of risk the agent is willing to take. If the safety critic indicates that the current state and action visited by the task policy is unsafe, the recovery policy will overwrite the task policy's actions, moving the agent back to safe regions of the state space. Both policies can be trained using any reinforcement learning algorithm, where $\pi_{\rm task}$ optimizes task reward and $\pi_{\rm rec}$ minimizes $Q^{\pi}_{\text{risk}}$.

\subsection{Meta-learning}
Consider a task distribution $p(\mathcal{M})$ where tasks are sampled via $\mathcal{M}_i\sim p(\mathcal{M})$. In the RL setting, each task corresponds to an MDP, all of which share the same state and action spaces but may have varying dynamics (e.g. varying controller impedance for a legged robot). The goal in this work is to learn risk measures that rapidly adapt to new environments, such as when a robot's actuator loses power and it is forced to compensate with only the remaining actuators. We will briefly discuss how functions can be initialized for rapid adaptation to new tasks by training on similar tasks.

Meta-learning learns a model explicitly optimized for adaptation to a new task from $p(\mathcal{M})$.
Let $\theta_{i}' = \theta - \alpha \nabla_{\theta} \mathcal{L}_{\mathcal{M}_i}(f_{\theta})$ be the parameters $\theta$ after a single gradient step from optimizing $\mathcal{L}_{\mathcal{M}_i}(f_{\theta})$.
Model-Agnostic Meta-Learning (MAML) \cite{finn2017modelagnostic} optimizes the following objective at meta-train time:
\begin{align}
\begin{split}
  \min_{\theta} \mathbb{E}_{\mathcal{M}_i \sim p(\mathcal{M})}\left[\mathcal{L}_{\mathcal{M}_i}(f_{\theta_{i}'})\right] = \min_{\theta} \mathbb{E}_{\mathcal{M}_i \sim p(\mathcal{M})}\left[\mathcal{L}_{\mathcal{M}_i} (f_{\theta - \alpha \nabla_{\theta} \mathcal{L}_{\mathcal{M}_i}(f_{\theta})})\right]
\end{split}
\end{align}

After meta-training, to quickly adapt to a new test environment, MAML computes a task-specific loss function from an unseen task and updates $\theta$ with several gradient steps.
\section{Problem Statement}
\label{sec:PS}
We consider the offline meta-reinforcement learning problem setting introduced in~\cite{mitchell2020offline, dorfman2021offline}, in which the objective is to leverage offline data from a number of different tasks to rapidly adapt to an unseen task at test-time. We consider an instantiation of this setting in which tasks correspond to CMDPs $\{\mathcal{M}_i\}_{i=1}^{N}$, each with different system dynamics $p_i(s' | s, a)$, but which otherwise share all other MDP parameters, including the same state and action spaces and constraint function. Here the agent is not allowed to directly interact with any environment at meta-train time or meta-test time, but is only provided with a fixed offline dataset of transitions from environments. This setting is particularly applicable to the safe reinforcement learning setting, where direct environmental interaction can be risky, but there may be accident logs from prior robot deployments in various settings. 
We formalize the problem of learning about constraints in the environment in the context of offline meta-reinforcement learning, in which the agent is provided with offline data from $N_\textrm{train}$ training environments $\{\mathcal{M}^{\textrm{train}}_i\}_{i=1}^{N_\text{train}}$
with varying system dynamics and must rapidly adapt to being safe in a new environment $\mathcal{M}^{\textrm{test}}$ with unseen system dynamics. The intuition is that when dynamics change, the states which violate constraints remain the same, but the behaviors that lead to these states may be very different. Thus, we consider the problem of using data from a number of training environments to optimize the safe RL objective in Equation~\ref{eq:safe_rl_obj}. 

We assume that the agent is provided with a set of $N_\textrm{train}$ datasets of offline transitions $\mathcal{D}^{\textrm{train}} = \{\mathcal{D}^{\textrm{train}}_i\}_{i=1}^{N_\textrm{train}}$ from training environments with different dynamics in addition to a small dataset $\mathcal{D}^{\textrm{test}}$ of offline transitions from the test environment $\mathcal{M}^{\textrm{test}}$, in which the agent is to be deployed. The agent's objective is to leverage this data to optimize the safe RL objective in Equation~\ref{eq:safe_rl_obj} in MDP $\mathcal{M}^{\textrm{test}}$ by learning some task $\tau$ in MDP $\mathcal{M}^{\textrm{test}}$ while minimizing constraint violations.
\section{\algname{}}
\label{sec:impl}
\begin{figure*}[t!]
    \centering
    \includegraphics[width=0.9\textwidth]{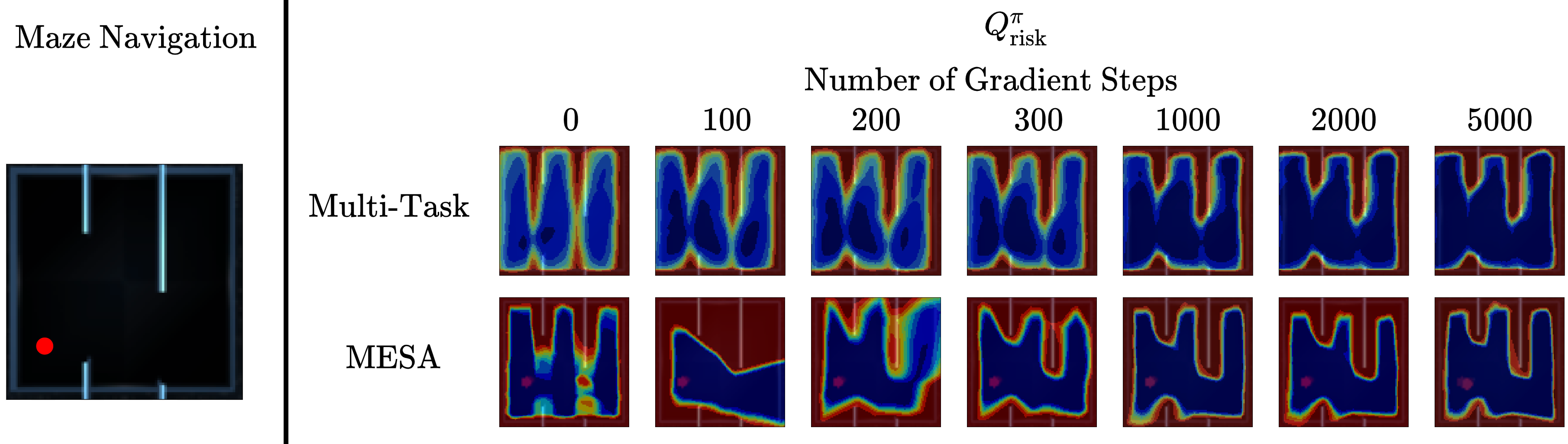}
    \caption{\textbf{Safety Critic Adaptation Visualizations: }\small For purposes of illustration, we evaluate \algabbr{} and a Multi-Task learning comparison on a simple Maze Navigation task (left) from~\cite{recovery-rl} in which the objective is for the agent (the red dot) to navigate from a random point in the left column to the middle of the right column without colliding into any of the Maze walls or boundaries. Environments are sampled by changing the gaps in the walls (parameterized by $w_1, w_2 \sim \unif(-0.1, 0.1)$), leading to significant changes in which behaviors are safe. On the left, we show heatmaps of the learned safety critic $Q^\pi_{\rm risk}$ when it is adapted to a new Maze with unseen wall gaps for the Multi-Task comparison (top) and \algabbr{} (bottom). Here bluer colors denote low probability of constraint violation while redder colors denote a higher probability, and the labels above the heatmaps indicate the number of gradient steps used for adaptation on $\mathcal{D}^\textrm{test}$. The Multi-Task learning comparison, which aggregates data from all environments to learn the safety critic and does not explicitly optimize for adaptation, is much slower to adapt while \algabbr{} is able to leverage its learned prior to rapidly adapt to the new gap positions.}
    \label{fig:safety-critic}
 \end{figure*}

We introduce \algname{}, a 3-phase procedure to optimize the objective in Section~\ref{sec:PS}. First, \algabbr{} uses datasets of offline transitions from the training environments to meta-learn a safety critic optimized for rapid adaptation (Section~\ref{subsec:meta-learning}). Then, we discuss how \algabbr{} adapts its meta-learned safety critic using a dataset of offline transitions from the test environment (Section~\ref{subsec:adaptation}). This same dataset is also used to learn a recovery policy, which is trained to descend the safety critic and prevent the agent from visiting unsafe states as in~\citet{recovery-rl}, but we note that the learned safety critic can also be used in conjunction with other safe RL algorithms such as those from~\citet{learn-to-be-safe, cql-safety}. Finally, the meta-learned safety critic and recovery policy are used and updated online when learning some downstream task $\tau$ in the testing environment (Section~\ref{subsec:safe-rl}). The full algorithm is summarized in Algorithm~\ref{alg:main} and Figure~\ref{fig:method}. An illustration of the safety critic adaptation procedure is shown in Figure~\ref{fig:safety-critic}.

\subsection{\bf Phase 1, Meta-Learning $Q^\pi_{\rm risk}$} 
\label{subsec:meta-learning}
Given offline transitions from $N_{\textrm{train}}$ training environments, $\{\mathcal{D}^{\textrm{train}}_i\}_{i=1}^{N_\textrm{train}}$, we meta-learn the safety critic $Q^\pi_{\psi, \rm risk}$, with parameters $\psi$, using Model-agnostic Meta Learning~\cite{finn2017modelagnostic}. We utilize the same safety critic loss function from~\cite{recovery-rl}. The recovery policy is not trained with a MAML-style objective. Similar to the actor's loss function in DDPG~\cite{DDPG}, the recovery policy, parameterized by $\omega$, aims to minimize the safety critic value for input state $s_t$:
\begin{align*}
\label{eq:actor_obj}
    & \mathcal{L}_{\pi_{\rm rec}}(\omega, s_t) = Q^{\pi}_{\psi, \rm risk}(s_t, \pi_{\omega, \text{rec}}(\cdot|s_t)).
\end{align*} 

\subsection{\bf Phase 2, Test Time Adaptation} 
\label{subsec:adaptation}
A previously unseen test environment $\mathcal{M}^{\rm test}$ is sampled from task distribution $p(\mathcal{M})$ and the agent is supplied with a dataset of offline transitions $\mathcal{D}_{\textrm{test}}$, which is \textbf{10-100x} smaller than the training datasets. We then
perform $M$ gradient steps with respect to $\mathcal{L}_{\text{risk}}(\psi, s)$ (in Section \ref{subsec:safetycriticintro}) and $\mathcal{L}_{\pi_{\rm rec}}(\omega, s)$ over $\mathcal{D}_{\textrm{test}}$ to rapidly adapt safety critic $Q^\pi_{\psi, \rm risk}$ and train recovery policy $\pi_{\omega, \rm rec}$

Note that the learned $Q^\pi_{\psi, \rm risk}$ is initially calibrated with the policy used for data collection in the meta-training environments. Since these datasets largely consist of constraint violations, the resulting $Q^\pi_{\psi, \rm risk}$ serves as a pessimistic initialization for online learning of some downstream task $\tau$. This is a desirable property, as $Q^\pi_{\psi, \rm risk}$ will initially prevent constraint violations, and then become increasingly less pessimistic during online exploration when calibrated with the task policy for task $\tau$.

\subsection{\bf Phase 3, Using $Q^\pi_{\rm risk}$ and $\pi_{\rm rec}$ for Safe RL} 
\label{subsec:safe-rl}
We initialize the safety critic and recovery policy with the adapted $Q^\pi_{\psi, \rm risk}$ and $\pi_{\omega, \textrm{rec}}$ when learning a task $\tau$ in the test environment. Since the safety critic is learned offline in a task-agnostic way, we can flexibly utilize the meta-learned safety critic and recovery policy to learn a previously unknown task $\tau$ in the test environment. As in Recovery RL~\cite{recovery-rl}, both $Q^\pi_{\psi, \rm risk}$ and $\pi_{\omega, \textrm{rec}}$ are updated online through interaction with the environment so that they are calibrated with the learned task policy for $\tau$.

\section{Experiments}
\begin{figure*}[t!]
    \centering
    \begin{subfigure}[t]{0.19\textwidth}
        \centering
        \includegraphics[width=\textwidth]{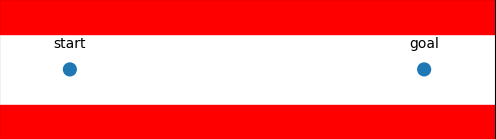}
        \caption{Navigation 1}
        \label{fig:pointbot-0}
    \end{subfigure}
    ~ 
    \begin{subfigure}[t]{0.19\textwidth}
        \centering
        \includegraphics[width=\textwidth]{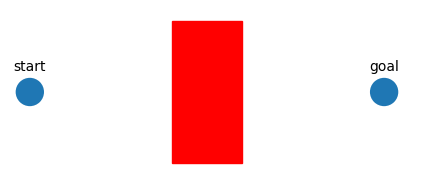}
    \caption{Navigation 2}
        \label{fig:pointbot-1}
    \end{subfigure}
    ~ 
    \begin{subfigure}[t]{0.18\textwidth}
        \centering
        \includegraphics[width=\textwidth]{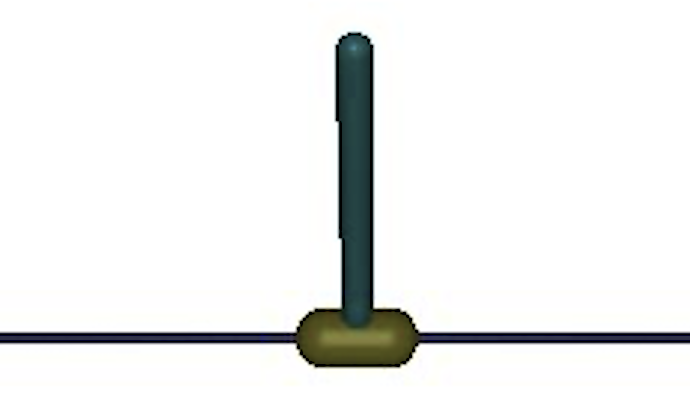}
        \caption{Cartpole Length}
        \label{fig:cartpole-length}
    \end{subfigure}
    ~ 
    \begin{subfigure}[t]{0.20\textwidth}
        \centering
        \includegraphics[width=\textwidth]{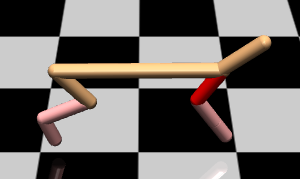}
        \caption{HalfCheetah Disabled}
        \label{fig:halfcheetah-disabled}
    \end{subfigure}
    ~ 
    \begin{subfigure}[t]{0.15\textwidth}
        \centering
        \includegraphics[width=\textwidth]{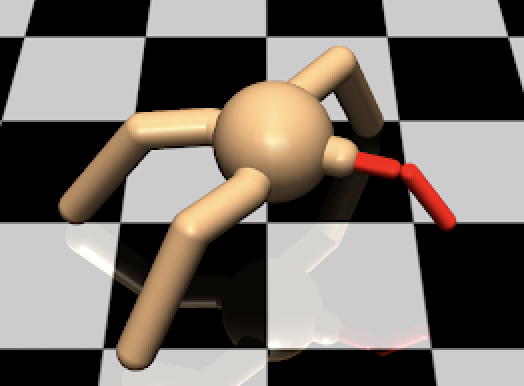}
        \caption{Ant Disabled}
        \label{fig:ant-disabled}
    \end{subfigure}
    
    \caption{\small \textbf{Simulation Domains:} We evaluate \agentname{} on a set of 2D navigation and locomotion tasks in simulation. In Navigation 1 and Navigation 2, the agent learns to navigate from a beginning position to the goal while avoiding the obstacles (red walls). In the Cartpole-Length task, the goal is to keep the pole balanced on the cart while minimizing the number of times the pole falls beneath the rail or moves off the rail. Lastly, in the HalfCheetah-Disabled and Ant-Disabled tasks, the objective is to learn how to move forwards while minimizing the number of collisions with the ground of the head (HalfCheetah) or torso (Ant) during training.}
	\label{fig:domains}
 \end{figure*}

\label{sec:exps}
We study the degree to which \algabbr{} can leverage offline data from environments with different dynamics to quickly learn safety in a new test domain with modified, previously unseen dynamics via a small amount of experience in the new domain. To do this, we compare \algabbr{} with prior safe reinforcement learning algorithms and study the degree to which they can limit constraint violations when learning in a perturbed test environment with previously unseen dynamics. \algname{} and all comparisons are built on top of the Soft Actor Critic (SAC) algorithm from~\citet{SAC}. 
\noindent {\bf Comparisons:} 
We compare \algabbr{} with the following algorithms: \textbf{Unconstrained: } A soft actor critic agent which only optimizes for task rewards and ignores constraints; \textbf{Recovery RL (RRL): } Uses data only from $\mathcal{D}_\mathrm{test}$ to learn $Q^{\pi}_{\text{risk}}$ and then uses $Q^{\pi}_{\text{risk}}$ in conjunction with the Recovery RL algorithm~\cite{recovery-rl}; \textbf{Multi-Task Learning (Multi-Task): } Learns $Q^{\pi}_{\text{risk}}$ from a combination of all data from both the training datasets $\{\mathcal{D}_i\}_{i=1}^{N_\textrm{train}}$ in phase 1 and then adapts in phase 2 using gradient steps on only the test dataset $\mathcal{D}_\mathrm{test}$. In phase 3, Multi-Task uses the learned $Q^{\pi}_{\text{risk}}$ in conjunction with the Recovery RL algorithm~\cite{recovery-rl} as in \algabbr{} and the RRL comparison; \textbf{CARL:} A prior safe meta-reinforcement learning algorithm which learns a dynamics model and safety indicator function through interaction with number of source environments and uses the uncertainty of the learned dynamics models to adapt to a target environment with previously unknown dynamics in a risk-averse manner; \textbf{CARL-Offline:} A modification of CARL which only provides CARL with offline datasets from the source environments, consistent with the offline meta-RL setting we consider in this work.

The comparison to Unconstrained allows us to evaluate the effect of reasoning about constraints at all. The comparison to Recovery RL allows us to understand whether offline data from different environments enables \algabbr{} to learn about constraints in the test environment. The comparison to the Multi-Task Learning algorithm allows us to evaluate the benefits of specifically leveraging meta-learning to quickly adapt learned risk measures. The comparisons to CARL and CARL-Offline allow us to evaluate whether \algabbr{} can outperform prior work in safe meta-RL.

\noindent {\bf Experimental Procedure:} 
We evaluate \algabbr{} and comparisons on their ability to (1) efficiently learn some downstream task $\tau$ in the test environment (2) while satisfying constraints. We report learning curves and cumulative constraint violations for all algorithms to see if \algabbr{} can leverage prior experience to safely adapt in the test environment. Episodes are terminated upon a constraint violation, making learning about constraints critical for safely learning in the test environment. We report average performance over 5 random seeds with standard error shading for all learning curves.

\begin{wrapfigure}{t!}{0.5\textwidth}
    \includegraphics[width=0.5\textwidth]{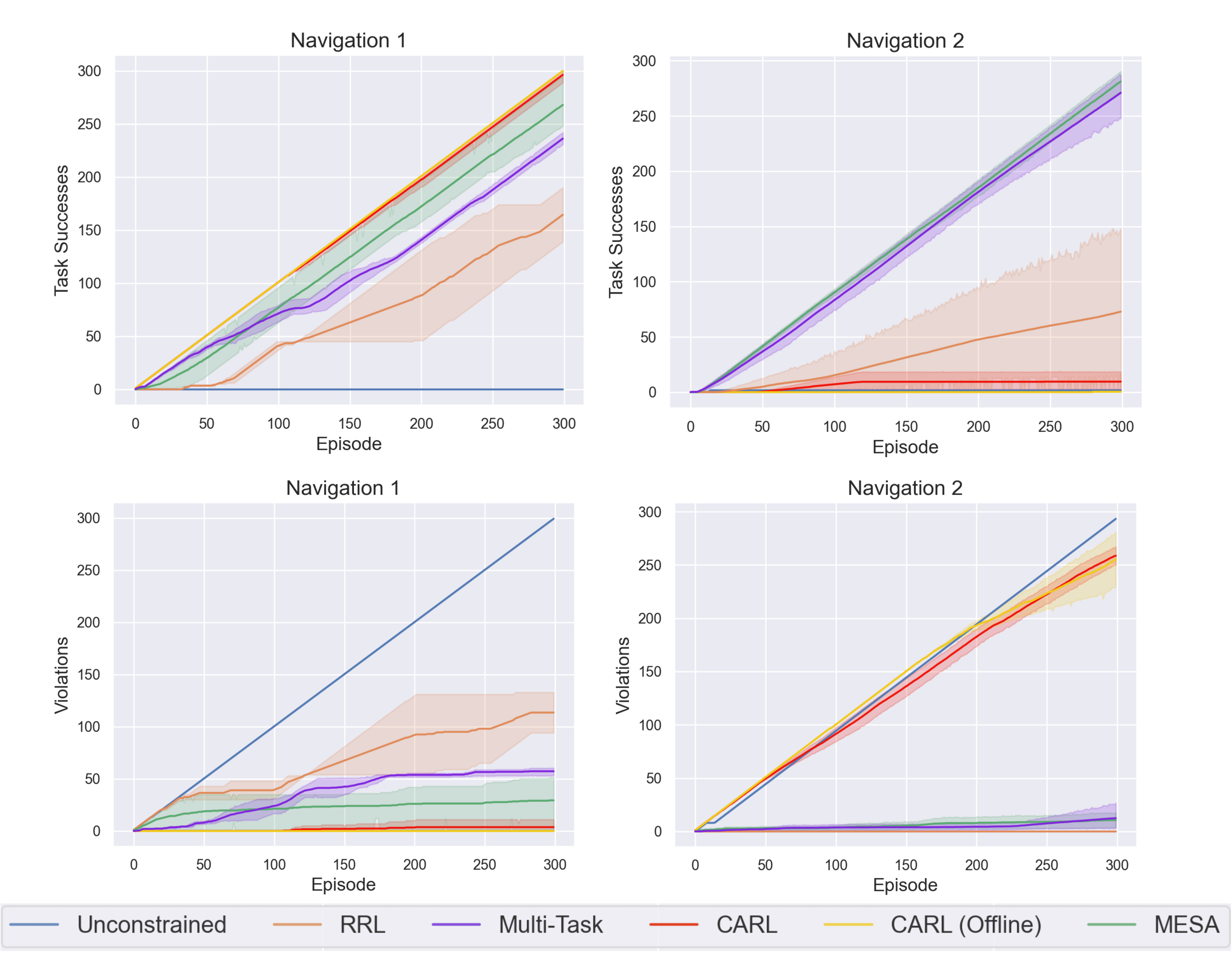}
    \caption{\small \textbf{Navigation Results: Top: Learning Curves (Phase 3).} \algabbr{} is able to achieve similar task success compared to prior algorithms on bot domains. \textbf{Bottom: Cumulative Constraint Violations (Phase 3). }Here, we find that \algabbr{} achieves fewer constraint violations than most comparisons, but find that the Multi-Task comparison also performs well on these environments.}
	\label{fig:navigation_results}
	\vspace{-3mm}
 \end{wrapfigure}

\noindent {\bf Domains:} 
We evaluate \algabbr{} and comparisons on 5 simulation domains which are illustrated in Figure~\ref{fig:domains}. All domains we study have the property that the changes in the dynamics are not immediately observable in the agent's observation, motivating learning how to be safe from interaction experience when dynamics change. This is common in various practical settings, such as a robot with worn out joints or sudden power loss in a legged locomotion system. We first consider two 2D navigation domains from~\cite{recovery-rl} in which the agent must navigate between a start set and goal set without colliding into red obstacles in a system with linear Gaussian dynamics. 
The environment distribution for both domains is defined by varying the 
coefficients of the $A$ and $B$ matrices in the transition dynamics function where $s_{t+1} = A \cdot s_{t} + B \cdot a_t + \epsilon $, where $\epsilon \sim \mathcal{N}(0, \sigma^2 I)$.

We then consider a cartpole task (Cartpole-Length) in which the agent must balance the cartpole system without letting the pole fall below the cart. Here environments are sampled by varying the length of the pole, where pole lengths for the training environments are sampled from $\unif(0.4, 0.8)$ and the test environment corresponds to a pole of length $1$. We also consider two legged locomotion tasks, HalfCheetah-Disabled and Ant-Disabled, in which the agent is rewarded for running as fast as possible, but violates constraints given a collision of the head with the floor or torso with the floor for the HalfCheetah-Disabled and Ant-Disabled tasks respectively. For both HalfCheetah-Disabled and Ant-Disabled, environments are sampled by choosing a specific joint and simulating a loss of power (power loss corresponds to always providing zero motor torque to the joint), resulting in significantly different dynamics across environments. The Cartpole-Length and HalfCheetah-Disabled tasks are adapted from~\cite{cautious-adaptation} while the Ant-Disabled task is from~\cite{nagabandi2019learning}. 
\subsection{Data Collection}
\label{subsec:collection}
For the navigation environments, offline datasets are collected via a random policy where the episode \textit{does not terminate} upon constraint violation.  We collect a total of 20-25 datasets for each of the sampled training environments, with each dataset consisting of 10000 transitions (680 and 1200 violations in Navigation 1 and Navigation 2 respectively), similar to that of \cite{recovery-rl}. However, the dataset in the test environment is \textbf{50-100x} smaller than each training task dataset ($\sim$100, 200 transitions with 15, 36 violations respectively).
\begin{figure*}[!t]
    \vspace{-3mm}
    \centering
    \includegraphics[width=0.95\textwidth]{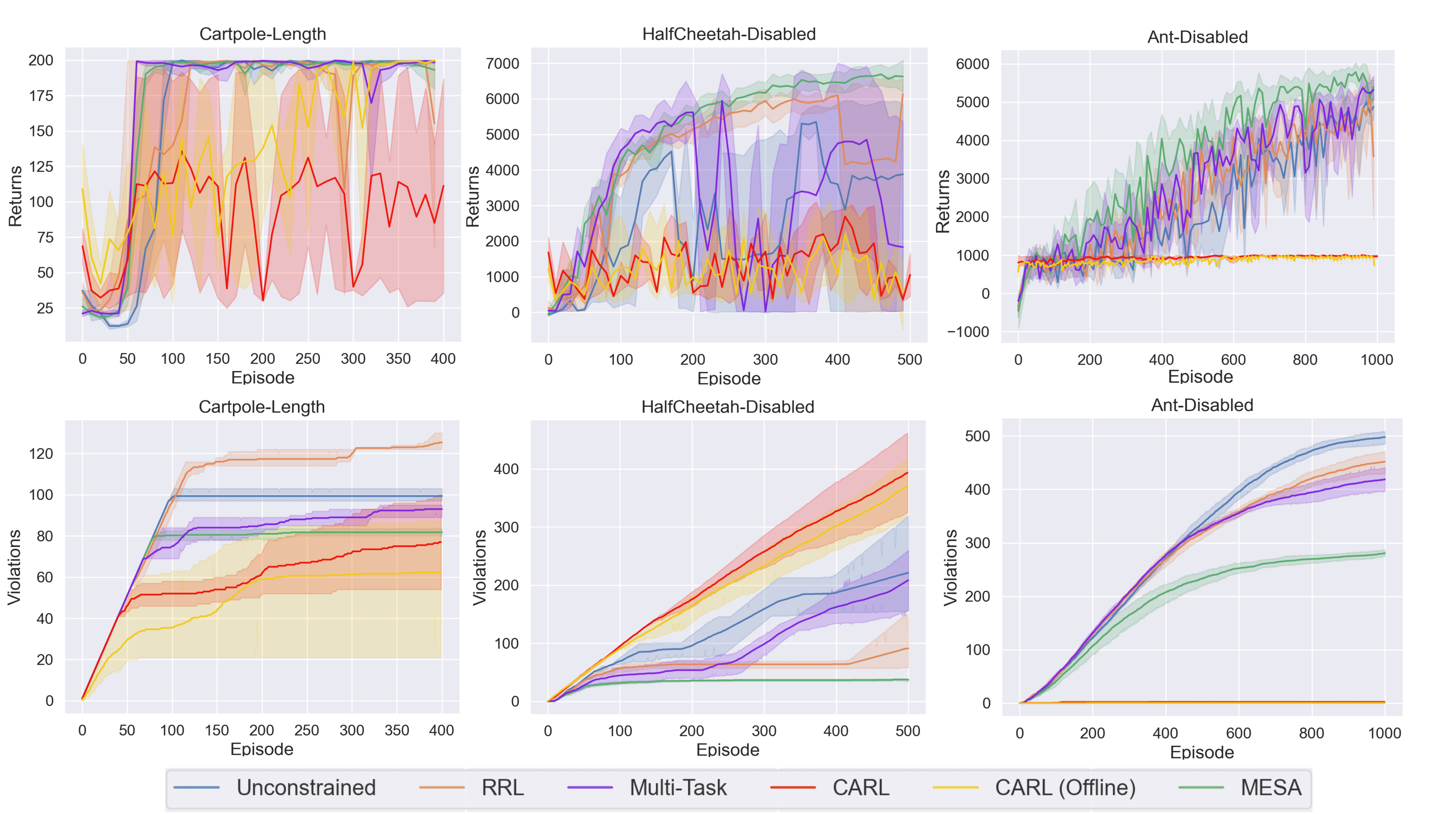}
    \caption{\small \textbf{Locomotion Results:} \textbf{Top: Learning Curves (Phase 3).} \algabbr{} achieves similar task performance as the best comparison algorithm, indicating that \algabbr{} is able to effectively learn in a test environment with previously unseen dynamics. \textbf{Bottom: Cumulative Constraint Violations (Phase 3).} \algabbr{} violates constraints less often than comparisons, with this difference being most significant on the HalfCheetah-Disabled and Ant-Disabled tasks. This suggests that \algabbr{} is able to effectively leverage its prior experiences across environments with different dynamics to rapidly adapt its risk measure to the test environment.}
    \label{fig:locomotion-results}
 	\vspace{-6mm}
 \end{figure*}

Similarly, for locomotion environments, the datasets from the test environment are collected via a random policy rollout, where the episode does not terminate early upon constraint violations. To collect datasets from the training environments, we train SAC on each of the training environments and log the replay buffer from an intermediate checkpoint. For the HalfCheetah-Disabled and Ant-Disabled tasks, we collect 4 and 3 training datasets of 400 episodes (on average $\sim$400K transitions with 14K and 113K violations) respectively. The dataset from the testing environment consists of 40K transitions (2.4K, and 11.2K violations for HalfCheetah, Ant), which is \textbf{10x} smaller than before. For the Cartpole-Length task, 20 training datasets are generated, with each containing 200 episodes of data ($\sim$20K timesteps with 4.5K violations). The dataset from the testing environment contains 1K transitions (with 200 violations), which is \textbf{20x} smaller than before. 

\subsection{Results}
\label{subsec:results}
\textbf{Navigation Results}: We evaluate the performance of \algabbr{} and comparisons in Figure~\ref{fig:navigation_results}. Unconstrained SAC performs poorly as it no mechanism to reason about constraints and thus collides frequently and is unable to learn the task. \algabbr{} violates constraints less often than the Multi-Task comparison, but the performance gap is somewhat small in these environments. We hypothesize that this is because in the Navigation environments, particularly Navigation 2, the space of safe behaviors does not change significantly as a function of the system dynamics, making it possible for the Multi-Task comparison to achieve strong performance by simply learning the safety critic jointly on a buffer of all collected data. CARL and CARL-Offline baselines perform the best in the Navigation 1 environment but are unable to make much progress in Navigation 2.

\textbf{Locomotion Results:} \algabbr{} significantly outperforms prior methods on the HalfCheetah-Disabled and Ant-Disabled, while achieving comparable performance on the Cartpole task (Figure~\ref{fig:locomotion-results}). We hypothesize that in the HalfCheetah-Disabled and Ant-Disabled tasks, the different training environments are sufficiently different in their dynamics that a safety critic and recovery policy trained jointly on all of them is unable to accurately represent the boundaries between safe and unsafe states. Thus, when adapting to an environment with unseen dynamics, the space of safe behaviors may be so different than in the training environments that the Multi-Task comparison cannot easily adapt. \algabbr{} mitigates this by explicitly optimizing the safety critic for rapid adaptation to different dynamics. In addition, CARL and CARL-Offline make little task progress in the HalfCheetah and Ant Disabled domain and, as a result, are able to generally satisfy constraints. The sharp decline in performance is likely due to the planning algorithm that CARL utilizes for optimization over learned dynamics.

\section{Ablations}
\begin{figure*}[t!]
    \centering
    \begin{subfigure}[t]{0.49\textwidth}
        \centering
        \includegraphics[width=\textwidth]{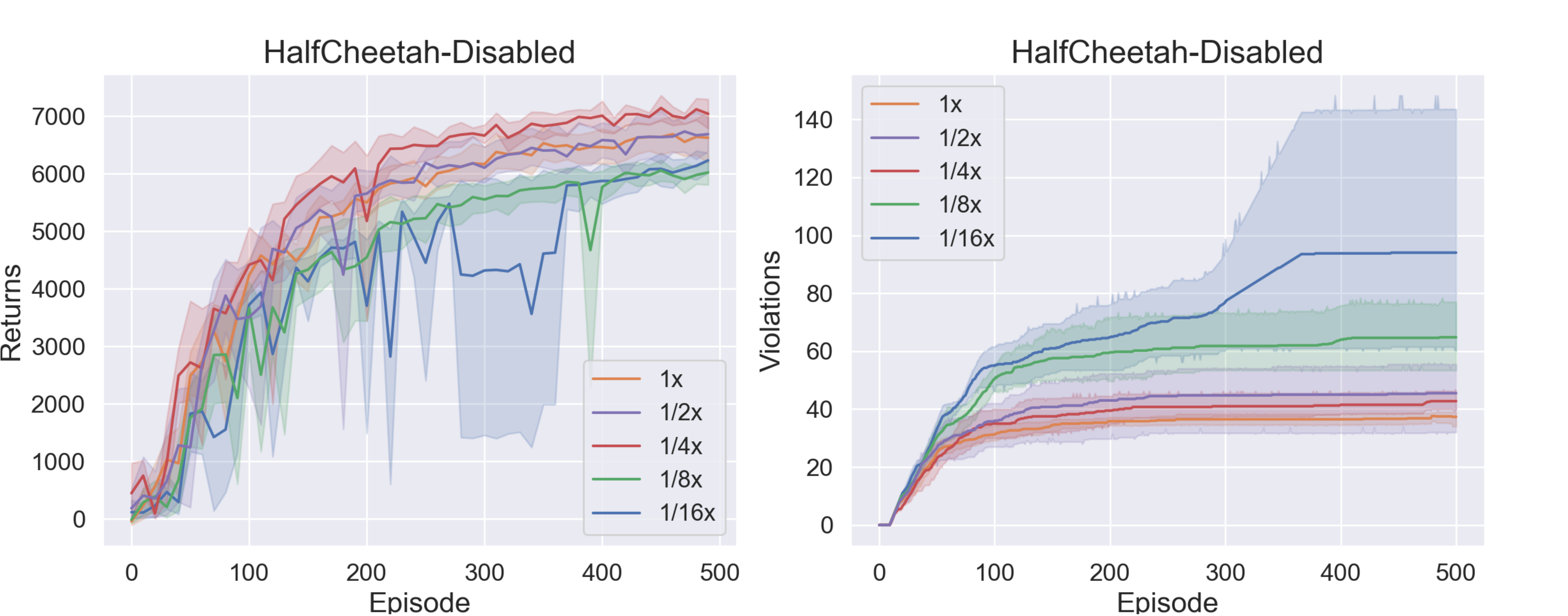}
        \caption{Varying Test Dataset Sizes}
        \label{fig:ablations-1}
    \end{subfigure}
    ~ 
    \begin{subfigure}[t]{0.49\textwidth}
        \centering
        \includegraphics[width=\textwidth]{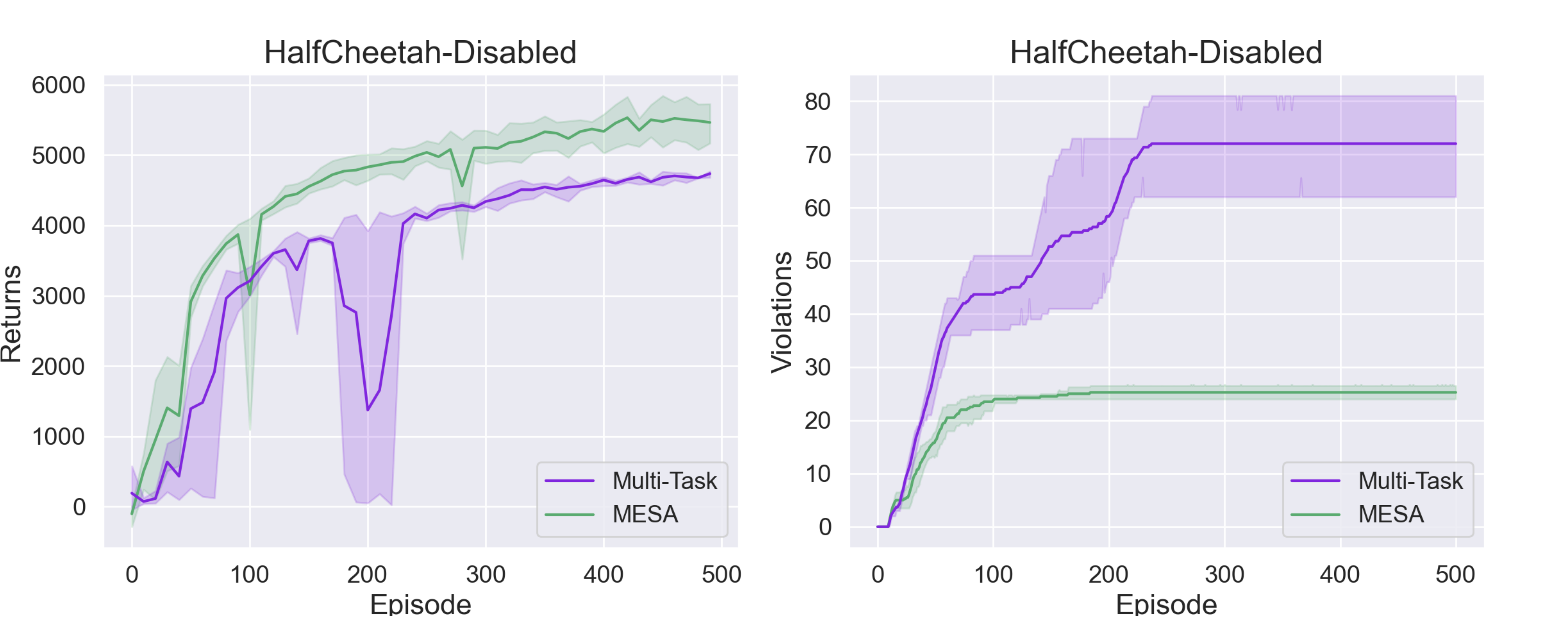}
    \caption{Test Task Generalization: Partial Joint Failures}
        \label{fig:ablations-2}
    \end{subfigure}
    \caption{\small \textbf{Ablation}: \textbf{Sensitivity to Test Dataset Size: }In Figure~\ref{fig:ablations-1}, we investigate the sensitivity of \algabbr{} to the number of transitions in the test dataset used for adapting $Q^\pi_\textrm{risk}$ for HalfCheetah-Disabled. We find that even with a test dataset 4 times smaller than used in the experiments in Section~\ref{sec:exps}, \algabbr{} does not experience much degradation in performance. However, further reductions in the test dataset size make it difficult for \algabbr{} to learn a sufficiently accurate safety critic in the test environment, leading to more significant drops in performance. \textbf{Generalization to More Different Test Environment Dynamics: }In Figure~\ref{fig:ablations-2}, we investigate \algabbr{}'s and Multi-Task's generalization to partial joint failures in the HalfCheetah-Disabled task, where the training sets are kept the same as described in Section \ref{sec:exps}. We find that \algabbr{} is able to significantly reduce the number of constraint violations compared to the Multi-Task comparison while also achieving superior task performance, suggesting that as differences in system dynamics increase between the training and testing environments, \algabbr{} is able to more effectively adapt risk measures across the environments.}
	\label{fig:ablations}
	\vspace{-2mm}
 \end{figure*}
\label{sec:extra-exps}
In ablations, we seek to answer the following questions: (1) how small can the dataset from the test environment be for \algabbr{} to safely adapt to new test environments? and (2) how well can \algabbr{} generalize to environments consisting of more significantly different dynamics (e.g. partial joint failures when only trained on datasets with examples of full joint failures)?

\subsection{Test Dataset Size}
We first investigate the sensitivty of \algabbr{} to the size of the test dataset. Figure~\ref{fig:ablations-1}, we study performance when the test dataset is 1x, 1/2x, 1/4x, 1/8x, and 1/16x the size of the test dataset (40K transitions) used for the HalfCheetah-Disabled results reported in Section~\ref{sec:exps}. We find that \algabbr{} can do well when given a test dataset 1/4 the size of the original test dataset (10K transitions, which is ~10 episodes of environment interaction). This suggests that the test size dataset can be up to \textbf{40x} smaller than the training dataset sizes without significant drop in performance. We find that when the test dataset is reduced to 1/8 and 1/16 the size of the original test dataset, \algabbr{} exhibits degrading performance, as the safety critic has insufficient data to learn about constraints in the test environment.

\subsection{Test Environment Generalization}
Here we study how \algabbr{} performs when the test environments have more significantly different dynamics from those seen during training. To evaluate this, we consider the HalfCheetah-Disabled task, and train \algabbr{} using the same training datasets considered in Section~\ref{sec:exps}, in which specific joints are selected to lose power. However, at test time, we evaluate \algabbr{} on a setting with partial power losses to joints, in which the maximum applicable power to certain joints is set to some $k$ percent of the original maximum power, where $k \in \unif(0.5, 0.95)$. This is analogous to partial subsystem failures that can occur in real-world robotic systems. In, Figure~\ref{fig:ablations-2}, we find that \algabbr{} achieves superior performance compared to the the Multi-Task comparison in terms of both task performance and constraint violations during training. This suggests that \algabbr{} could rapidly learn to be safe even with system dynamics that are out of the meta-training environment distribution.
\section{Conclusion}
\label{sec:conclusion}
We formulate safe reinforcement learning as an offline meta-reinforcement learning problem and motivate how learning from offline datasets of unsafe behaviors in previous environments can provide a scalable and compelling way to learn tasks safely in new environments with unobserved change in system dynamics. We then present \algname, a new algorithm for learning a risk measure which can transfer knowledge about safety across environments with different dynamics. Results in simulation experiments suggest that \algabbr{} is able to achieve strong performance across $5$ different robotic simulation domains and is able to effectively adapt to test environments with previously unseen dynamics.

\clearpage
\section{Acknowledgments}
\footnotesize
This research was performed at the AUTOLAB at UC Berkeley in affiliation with the Berkeley AI Research (BAIR) Lab, the Real-Time Intelligent Secure Execution (RISE) Lab, Google Brain Robotics, and the Stanford AI Research Lab. Authors were also supported by the SAIL-Toyota Research initiative, the Scalable Collaborative Human-Robot Learning (SCHooL) Project, the NSF National Robotics Initiative Award 1734633, and in part by donations from Google, Siemens, Amazon Robotics, Toyota Research Institute, and by equipment grants from NVidia. This article solely reflects the opinions and conclusions of its authors and not the views of the sponsors or their associated entities. A.B. and S.N. were supported by NSF GRFPs. We thank our colleagues who provided helpful feedback, code, and suggestions, in particular Daniel Brown and Ellen Novoseller.

\bibliographystyle{plainnat}
\bibliography{yummy}
\clearpage

\appendix
\section{Appendix}
\label{sec:appendix}

\subsection{Algorithm Description}

The full detail of the MESA algorithm is described in Algorithm ~\ref{alg:main}. In Phase 1, Offline Meta-Learning, the safety critic is updated with a MAML-style objective. In Phase 2, both the safety critic and recovery policy adapt to the test environment with a small offline test dataset. Finally, in Phase 3, the agent interacts with the test environment by using Recovery RL ~\cite{recovery-rl} to avoid constraint violations.
\begin{algorithm}
\caption{\algname}
\label{alg:main}
\begin{algorithmic}
\Require{Training datasets $\mathcal{D}^{\rm train} = \{\mathcal{D}^{\rm train}_i\}_{i=1}^{N_\textrm{train}}$, adaptation dataset $\mathcal{D}^\textrm{test}$, task horizon $H$, safety threshold $\epsilon_{\text{risk}}$}, safety critic step sizes $\alpha_1$ and $\alpha_2$, recovery policy step size $\beta$.
\For{$i \in \{1,\ldots N\}$}
    \Comment{Phase 1: Offline Meta-Learning}
    \For{$j \in \{1,\ldots K\}$}
            \State{$\textrm{Sample } \mathcal{D}^{\rm train}_{j} \sim \mathcal{D}^{\textrm{train}} $}
            \State {$\psi_{j}^{\prime} \leftarrow \psi- \alpha_1 \cdot \nabla_{\psi} \mathcal{L}_{\textrm{risk}}\left(\psi, \mathcal{D}^{\rm train}_{j}\right)$}
    \EndFor
    
    \For{$j \in \{1,\ldots K\}$}
            \State{$\textrm{Sample } \mathcal{D}^{\rm test}_{j} \sim \mathcal{D}^{\textrm{test}} $}
    \EndFor
    \State{$\psi \leftarrow \psi-\alpha_2 \cdot \sum_{j}\nabla_{\psi} \mathcal{L}_{\textrm{risk}}\left(\psi_{j}^{\prime}, \mathcal{D}_{j}^{\textrm{train}}\right)$}
    \State $ \omega \leftarrow \omega- \beta \cdot \nabla_{\omega} \mathcal{L}_{\pi_\textrm{rec}}\left(\omega, \mathcal{D}^{\textrm{test}}\right)$
\EndFor
\For{$i \in \{1,\ldots M\}$}
    \Comment{Phase 2: Test Time Adaptation}
    \State $ \psi \leftarrow \psi - \alpha_1 \cdot \nabla_{\psi} \mathcal{L}_{\textrm{risk}}\left(\psi, \mathcal{D}^{\textrm{test}}\right)$
    \State $ \omega \leftarrow \omega- \beta \cdot \nabla_{\omega} \mathcal{L}_{\pi_\textrm{rec}}\left(\omega, \mathcal{D}^{\textrm{test}}\right)$
\EndFor
\State{$\mathcal{D}^{\mathrm{task}} \leftarrow \emptyset$}
\While{$\text{not converged}$}
    \Comment{Phase 3: Recovery RL}
    \State $s_1 \sim \texttt{env.reset()}$
    \For{$t \in \{1,\ldots H\}$}
        \State{$a^\pi_t, a_t^{\textrm{rec}} \sim \pi_{\theta}(\cdot | s_{t}), \pi_\textrm{rec}(\cdot | s_{t})$}
        \If{$Q^{\pi}_{\textrm{risk}}(s_t, a_t) \leq \epsilon_{\textrm{risk}}$}
            \State{$a_t = a^\pi_t$}
        \Else
            \State{$a_t = a_t^{\textrm{rec}}$}
        \EndIf
        \State Execute $a_t$, observe $r_t$, $c_t$, and $s_{t+1}$
        \State{Add $(s_t, a_t^{\pi}, c_t, s_{t+1})$ to $\mathcal{D}^{\textrm{test}}$}
        \State{Add $(s_t, a_t, r_t, s_{t+1})$ to $\mathcal{D}^{\text{task}}$}

        \State{$ \theta \leftarrow \theta- \gamma \cdot \nabla_{\theta} \mathcal{L}_{\pi}\left(\theta, \mathcal{D}^{\textrm{task}}\right)$}
        \State{$ \psi \leftarrow \psi - \alpha_1 \cdot \nabla_{\psi} \mathcal{L}_{\textrm{safe}}\left(\psi, \mathcal{D}^{\textrm{test}}\right)$}
        \State{$ \omega \leftarrow \omega- \beta \cdot \nabla_{\omega} \mathcal{L}_{\pi_{\rm rec}}\left(\omega, \mathcal{D}^{\textrm{test}}\right)$}
        \If{$c_t$}
            \State End episode
        \EndIf
    \EndFor
\EndWhile
\end{algorithmic}
\end{algorithm}

\subsection{Hyperparameters for \algabbr{} and Comparisons}

\begin{table}
\centering
\begin{tabular}{l|cccc}
\toprule
{\sc Hyperparameters} & {\sc Unconstrained} & {\sc RRL} & {\sc Multi-Task} & {\sc \algabbr{}} \\
\midrule
\multicolumn{5}{ c }{\textbf{Phase 1}: Offline Training ($\mathcal{D}_\textrm{train}$)}\\
\midrule
Total Iterations & --- & --- & 10000 & 10000 \\
Inner Batch Size $|B_\textrm{in}|$ & --- & --- & --- & 256 \\
Outer Batch Size $|B_\textrm{out}|$ & --- & --- & --- & 256 \\
Inner Adaptation Steps & --- & --- & --- & 1 \\
Inner LR $\alpha_1$ & --- & --- & --- & 0.001 \\
Outer LR $\alpha_2$ & --- & --- & --- & 0.00001 \\ 
Task Batch Size $K$ & --- & --- & --- & 5 \\
Adam LR $\eta$ & --- & --- & 0.0003 & --- \\
Batch Size $B$ & --- & --- & 256 & --- \\
\midrule
\multicolumn{5}{ c }{\textbf{Phase 2}: Offline Finetuning ($\mathcal{D}_\textrm{test}$)}\\
\midrule
Total Iterations $M$ & --- & 10000 & 500 & 500 \\
Batch Size $B$ & --- & 256 & 256 & 256 \\
Adam LR & --- & 0.0003 & $\eta$ & $\alpha_1$ \\
\midrule
\multicolumn{5}{ c }{\textbf{Phase 3}: Online Finetuning} \\
\midrule
Adam LR & 0.0003 & 0.0003 & $\eta$ & $\alpha_1$ \\
Batch Size $B$  & 256 & 256 & 256 & 256 \\
Discount $\gamma$ & 0.99 & 0.99 & 0.99 & 0.99 \\
$ \gamma_\textrm{risk}$ & --- & 0.8 & 0.8 & 0.8 \\
$\epsilon_\textrm{risk}$ & --- & 0.1 & 0.1 & 0.1 \\

\bottomrule
\end{tabular}
\caption{Algorithm Hyperparameters.}
\label{table:algorithm_hyperparameters}
\end{table}

\begin{table}
\centering
\begin{tabular}{l|cccc}
\toprule
{\sc Hyperparameters} & {\sc Unconstrained} & {\sc RRL} & {\sc Multi-Task} & {\sc \algabbr{}} \\
\midrule
\multicolumn{5}{ c }{\textbf{Phase 2}: Offline Finetuning ($\mathcal{D}_\textrm{test}$)}\\
\midrule
Total Iterations $M$ & --- & 2000 & 100 & 100 \\
Batch Size $B$ & --- & 64 & 64 & 64 \\
\midrule
\multicolumn{5}{ c }{\textbf{Phase 3}: Online Finetuning} \\
\midrule
$ \gamma_\textrm{risk} $ (Navigation 2) & --- & 0.65 & 0.65 & 0.65 \\
$\epsilon_\textrm{risk}$ (Navigation 1)  & --- & 0.3 & 0.3 & 0.3 \\

\bottomrule
\end{tabular}
\caption{Navigation Hyperparameter Differences}
\label{table:navigation_hyperparameters}
\end{table}

\begin{table}
\centering
\begin{tabular}{l|cccc}
\toprule
{\sc Hyperparameters} & {\sc Unconstrained} & {\sc RRL} & {\sc Multi-Task} & {\sc \algabbr{}} \\
\midrule
\multicolumn{5}{ c }{\textbf{Phase 1}: Offline Training ($\mathcal{D}_\textrm{train}$)}\\
\midrule
Total Iterations & --- & --- & 15000 & 15000 \\
\bottomrule
\end{tabular}
\caption{HalfCheetah-Disabled Hyperparameter Differences.}
\label{table:hc_hyperparameters}
\end{table}

\begin{table}
\centering
\begin{tabular}{l|cccc}
\toprule
{\sc Hyperparameters} & {\sc Unconstrained} & {\sc RRL} & {\sc Multi-Task} & {\sc \algabbr{}} \\
\midrule
\multicolumn{5}{ c }{\textbf{Phase 1}: Offline Training ($\mathcal{D}_\textrm{train}$)}\\
\midrule
Total Iterations & --- & --- & 15000 & 15000 \\
\midrule
\multicolumn{5}{ c }{\textbf{Phase 3}: Online Finetuning} \\
\midrule
Risk Threshold $\epsilon_\textrm{risk}$ & --- & 0.3 & 0.3 & 0.3 \\
\bottomrule
\end{tabular}
\caption{Ant-Disabled Hyperparameter Differences.}
\label{table:ant_hyperparameters}
\end{table}

We report global hyperparameters shared across all algorithms in Table \ref{table:algorithm_hyperparameters} and additionally include domain specific hyperparameters in separate tables in Tables \ref{table:navigation_hyperparameters}, \ref{table:hc_hyperparameters}, and \ref{table:ant_hyperparameters}. We use the same base neural network architecture for the safety critic, recovery policy, actor for the task policy, and critic for the task policy. This base network is a fully connected network with 2 hidden layers each with 256 hidden units. For the task policy, we utilize the Soft Actor Critic algorithm from~\cite{SAC} and build on the implementation provided in~\cite{SAC_github}.

\subsection{Dataset Details}

To collect datasets from the training environments, we train SAC on each of the training environments and log the replay buffer from an intermediate checkpoint. For the HalfCheetah-Disabled and Ant-Disabled tasks, we collect 4 and 3 training datasets of 400 episodes (on average $\sim$400K transitions with 14K and 113K violations) respectively. The dataset from the testing environment consists of 40K transitions (2.4K, and 11.2K violations for HalfCheetah, Ant), which is \textbf{10x} smaller than before. For the Cartpole-Length task, 20 training datasets are generated, with each containing 200 episodes of data ($\sim$20K timesteps with 4.5K violations). The dataset from the testing environment contains 1K transitions (with 200 violations), which is \textbf{20x} smaller than before. 

\begin{table}[htb]
\centering
\begin{tabular}{l|ccc}
\toprule
{\sc Dataset} & {\sc $N_{\textrm{train}}$} & {\sc $|D_\textrm{train}|$} & {\sc $|D_\textrm{test}|$} \\
\midrule
Cartpole-Length & 20 & 20K & 1K \\ 
HalfCheetah-Disabled & 4 & 400K & 40K \\
Ant-Disabled & 3 & 400K & 40K \\
\bottomrule
\end{tabular}
\caption{Dataset Hyperparameters.}
\label{table:dataset_hypers}
\end{table}

For all environments, datasets are collected by an early-stopped SAC run, where the episode does not end on constraint violation. The testing dataset is collected by a randomly initialized policy. Each episode consists of 1000 timesteps.

\subsection{Related Work}
\label{sec:rw}
\subsubsection{Safe Reinforcement Learning}
There has been significant recent work on reinforcement learning algorithms which can satisfy safety constraints. We specifically focus on satisfying explicit state-space constraints in the environment and review prior literature which also considers this setting~\cite{safe-rl-survey}. Prior work has considered a number of methods for incorporating constraints into policy optimization for reinforcement learning, including trust region based methods~\cite{risk-constrained-RL, CPO}, optimizing a Lagrangian relaxation~\cite{RCPO, risk-sensitive-control,learn-to-be-safe, cql-safety}, drawing connections to Lyapunov theory~\cite{lyapunov-safety1, lyapunov-safety2, StabilitySafeMBRL}, anticipating violations with learned dynamics models~\cite{SAVED, abc-lmpc, LS3, cautious-adaptation}, using Gaussian processes to reason about uncertainty~\cite{safe-quadrotors1, safe-quadrotors2}, using recovery policies to shield the agent from constraint violations~\cite{ shielding, MPC-shielding, compound-RL, leave-no-trace, recovery-rl}, formal reachability analysis~\cite{safety-framework, HJ-reachability, inf-horizon, sampled-reachability, HJ-reach-avoid, safe-platooning}, or formal logic~\cite{shielding-formal-logic, chance-constrained-logic}.~\citet{cautious-adaptation} design a model-based RL algorithm which leverages unsafe data from a variety of training environments with different dynamics to predict whether the agent will encounter unsafe states and penalize its reward if this is the case. Unlike~\citet{cautious-adaptation}, we explicitly optimize for adaptation and decouple information about constraints from the reward function, making it possible to efficiently learn transferable notions of safety. Additionally, we learn a risk measure in a fully offline setting, and do not assume direct access to the training environments.

~\citet{learn-to-be-safe} introduce the idea of a safety critic, which estimates the discounted probability of constraint violation of the current policy given the current state and a proposed action.~\citet{learn-to-be-safe, cql-safety, recovery-rl} present $3$ different methods to utilize the learned safety critic for safe RL.
~\citet{learn-to-be-safe, recovery-rl} also leverage prior data from previous interactions to learn how to be safe. However, unlike these works, which assume that prior data is collected in an environment with the same dynamics as the test environment, \algabbr{} learns to leverage experience from a variety of environments with different dynamics in addition to a small amount of data from the test environment. This choice makes it possible to avoid excessive constraint violations in the test environment, in which constraint violations may be costly, by leveraging prior experience in safer environments or from accident logs from previous deployments.
\subsubsection{Meta Reinforcement Learning}
There is a rich literature \cite{schmidhuber1987, bengio1990learning, naik1992meta, thrun1998learning, Hochreiter01learningto} studying learning agents that can efficiently adapt to new tasks. In the context of reinforcement learning, this problem, termed \emph{meta-reinforcement learning} \cite{duan2016rl2, wang2017learning, finn2017modelagnostic}, aims to learn RL agents which can efficiently adapt their policies to new environments with unseen transition dynamics and rewards. A number of strategies exist to accomplish this such as recurrent or recursive policies \cite{duan2016rl2, wang2017learning, mishra2018simple}, gradient based optimization of policy parameters \cite{finn2017modelagnostic, houthooft2018evolved}, task inference \cite{rakelly2019efficient, humplik2019meta, fakoor2020metaqlearning}, or adapting dynamics models for model-based RL \cite{Saemundsson2018, nagabandi2019learning}.
One of the core challenges studied in many meta-RL works is efficient exploration \cite{stadie_2018_importance, rakelly2019efficient, zhang_learning_2020, liu2021decoupling}, since the agent needs to efficiently explore its new environment to identify the underlying task. Unlike all of these prior works, which focus on learning transferable policies, we focus on learning \textit{risk measures} which can be used to safely learn new tasks in a test environment with previously unseen dynamics. Additionally, we study learning these measures in the context of offline meta-RL, and learn from purely offline datasets of prior interactions in various environments with different dynamics.

The \emph{offline meta reinforcement learning} problem~\cite{mitchell2020offline, dorfman2021offline, li2021efficient} considers a setting in which the agent learns from a set of offline data from each training task, and adapts to the test environment conditioned only on a small set of offline transitions. Critically, this setting is particularly well suited to the problem of safe RL, because it has potential to enable an agent to be safe in an environment with previously unseen dynamics conditioned on a small set of experiences from that environment. In this work, we formalize safe reinforcement learning as an offline meta-RL problem and
present an algorithm to adapt \emph{a safety critic} 
to new environments and use this adapted safety critic for safe reinforcement learning.
One option for meta-learning for safe RL is using meta-learning for sim-to-real domain adaptation where data can be collected safely and at scale in simulated environments~\cite{meta-sim2real}. By contrast, \algabbr{} explicitly reasons about safety constraints in the environment to learn adaptable risk measures. Additionally, while prior work has also explored using meta-learning in the context of safe-RL~\cite{grbic2020safe}, specifically by learning a single safety filter which keeps policies adapted for different tasks safe, we instead adapt \emph{the risk measure itself} to unseen dynamics and fault structures.

\end{document}